\documentclass[journal=jcisd8,manuscript=article]{achemso}

\usepackage{chemformula} 
\usepackage[T1]{fontenc} 
\usepackage{amsmath}
\usepackage{amssymb}
\usepackage{multirow}
\usepackage{subcaption}
\usepackage[colorlinks=true, allcolors=blue]{hyperref}
\usepackage{graphicx}
\usepackage{makecell}
\usepackage[super]{natbib}

\title{Data-Driven Parametrization of Molecular Mechanics Force Fields for Expansive Chemical Space Coverage}

\author{Tianze Zheng}
\affiliation{ByteDance Research}
\email{zhengtianze@bytedance.com}
\author{Ailun Wang}
\affiliation{ByteDance Research}
\author{Xu Han}
\affiliation{ByteDance Research}
\author{Yu Xia}
\affiliation{ByteDance Research}
\author{Xingyuan Xu}
\affiliation{ByteDance Research}
\author{Jiawei Zhan}
\affiliation{ByteDance Research}
\alsoaffiliation{work done as intern at ByteDance Research}
\author{Yu Liu}
\affiliation{ByteDance Research}
\author{Yang Chen}
\affiliation{ByteDance Research}
\author{Zhi Wang}
\affiliation{ByteDance Research}
\author{Xiaojie Wu}
\affiliation{ByteDance Research}
\author{Sheng Gong}
\affiliation{ByteDance Research}
\author{Wen Yan}
\affiliation{ByteDance Research}
\email{wen.yan@bytedance.com}

\begin{document}

\begin{tocentry}
\includegraphics[width=\textwidth]{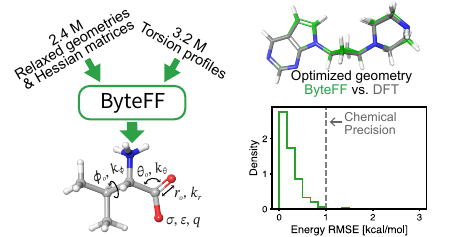}
\end{tocentry}

\begin{abstract}
    A force field is a critical component in molecular dynamics simulations for computational drug discovery. 
    It must achieve high accuracy within the constraints of molecular mechanics' (MM) limited functional forms, which offers high computational efficiency. 
    With the rapid expansion of synthetically accessible chemical space, traditional look-up table approaches face significant challenges. 
    In this study, we address this issue using a modern data-driven approach, developing ByteFF, an Amber-compatible force field for drug-like molecules.
    To create ByteFF, we generated an expansive and highly diverse molecular dataset at the B3LYP-D3(BJ)/DZVP level of theory.
    This dataset includes 2.4 million optimized molecular fragment geometries with analytical Hessian matrices, along with 3.2 million torsion profiles. 
    We then trained an edge-augmented, symmetry-preserving molecular graph neural network (GNN) on this dataset, employing a carefully optimized training strategy. 
    Our model predicts all bonded and non-bonded MM force field parameters for drug-like molecules simultaneously across a broad chemical space.
    ByteFF demonstrates state-of-the-art performance on various benchmark datasets, excelling in predicting relaxed geometries, torsional energy profiles, and conformational energies and forces. 
    Its exceptional accuracy and expansive chemical space coverage make ByteFF a valuable tool for multiple stages of computational drug discovery.
\end{abstract}

\section{Introduction}
Drug discovery involves identifying potential therapeutic candidates within the vast and intricate landscape known as chemical space, which encompasses all possible molecular structures. \cite{reymond2012exploring,warr2022exploration,caron2021steering} 
As a pivotal tool in this process, molecular dynamics (MD) simulations offer insights into dynamical behaviors and physical properties of molecules, as well as interactions in the molecular systems at an atomic level. \cite{liu2018molecular,de2016role,das2021accelerated,muegge2023recent,song2020evolution}
Central to the accuracy and reliability of these simulations is the force field, a mathematical model that describes the potential energy surface (PES) of the molecular system as a function of the positions of the atoms involved. \cite{wangDevelopmentTestingGeneral2004,harderOPLS3ForceField2016,roosOPLS3eExtendingForce2019,luOPLS4ImprovingForce2021}
Recent advances in synthetic chemistry and high-throughput screening technologies have significantly expanded the chemical space for drug candidates,\cite{buskes2020impact,guillemard2021late,brandenberg2020high,mohammad2021instadock,touret2020vitro} which necessitates the development of force fields that can provide accurate predictions of PES for diverse molecules in expansive chemical space.

Generally, force fields can be classified into two main categories: conventional molecular mechanics force fields (MMFFs) \cite{wangDevelopmentTestingGeneral2004,harderOPLS3ForceField2016,roosOPLS3eExtendingForce2019,luOPLS4ImprovingForce2021,qiuDevelopmentBenchmarkingOpen2021,boothroydDevelopmentBenchmarkingOpen2023}, which parameterize a fixed analytical form to approximate the energy landscape, and machine learning force fields (MLFFs) \cite{batatiaMACEHigherOrder2022,batznerEquivariantGraphNeural2022,rezaeeComparingANI2xANI1ccx2024,pelaezTorchMDNetFastNeural2024}, that aim to map the atomistic and molecular features and coordinates to the PES using neural networks without being limited by the fixed functional forms. 
Most conventional MMFFs, such as Amber\cite{hornak2006comparison,maier2015ff14sb}, GAFF\cite{wangDevelopmentTestingGeneral2004} and OPLS\cite{rizzo1999opls,roosOPLS3eExtendingForce2019,luOPLS4ImprovingForce2021}, describe the molecular PES by decomposing it into various degrees of freedom, including bonded (i.e., bonds, angles, and torsions) and non-bonded interactions (i.e., electrostatics and dispersion). 
These conventional MMFFs benefit from the computational efficiency of these terms, while suffering inaccuracies due to the inherent approximation, especially when non-pairwise additivity of non-bonded interactions are of significant importance.
Emerging recently, MLFFs have shown great promise for modeling PES due to their ability to capture subtle interactions and complex behaviors that may be overlooked or oversimplified by classical models. \cite{batatiaMACEHigherOrder2022,batznerEquivariantGraphNeural2022,rezaeeComparingANI2xANI1ccx2024,pelaezTorchMDNetFastNeural2024} 
Despite their outstanding accuracies, several drawbacks limit their applications in drug discovery. 
Owing to the complexity of the machine learning models involved, the computational efficiency of MLFFs is relatively low. 
Meanwhile, the amount of data required to train an effective MLFF is extremely large, which imposes constraints on their ability to cover the chemical space comprehensively.
Consequently, conventional MMFFs remain the most reliable and commonly used tool for MD simulations involving biological systems to this day. \cite{song2020evolution,muegge2023recent}

In the past few years, several efforts have been made to improve the quality of MMFFs for predicting the PES of small molecules. 
Following the traditional look-up table approach, OPLS3e increased the number of torsion types to 146,669 to enhance accuracy and expand chemical space coverage. \cite{roosOPLS3eExtendingForce2019}
Furthermore, OPLS3e and its successor OPLS4 were empowered with FFBuilder \cite{roosOPLS3eExtendingForce2019} to refine the torsion terms for molecules beyond the coverage of the pre-determined torsion list.
OpenFF\cite{wangOpenForceField2024a,qiuDevelopmentBenchmarkingOpen2021,boothroydDevelopmentBenchmarkingOpen2023} took a different approach by utilizing SMIRKS patterns to describe the chemical environment of both bonded and Lennard-Jones terms. 
However, these discrete descriptions of the chemical environment have inherent limitations that hamper the transferability and scalability of these force fields.
To tackle this problem, Espaloma\cite{wangEndtoendDifferentiableConstruction2022,takabaMachinelearnedMolecularMechanics2024} introduced a novel end-to-end workflow where the MMFF parameters are predicted by graph neural networks (GNN), opening new avenues for the advancement of MMFFs. \cite{chenAdvancingForceFields2024}
Despite promising results, these early attempts are constrained by the ML techniques and the training data, where significant improvements can be achieved. 

In this work, we propose ByteFF, a data-driven MMFF trained on a large-scale, high-diversity, and high-quality quantum mechanics (QM) dataset with sophisticated ML techniques. 
ByteFF is designed to leverage both atom and bond features using a state-of-the-art GNN model, while preserving molecular symmetry.
To construct the dataset for training ByteFF, we employ novel fragmentation methods and a rigorous QM calculation workflow, generating 2.4 million optimized molecular fragment geometries with Hessian matrices, along with 3.2 million torsion profiles.
Additionally, we introduce a differentiable partial Hessian loss and an iterative optimization-and-training procedure to effectively train ByteFF on the dataset.
Finally, we demonstrate the performance of ByteFF on various benchmarks, showing its expansive chemical space coverage and exceptional accuracy on intra-molecular conformational PES.

\section{Methods}
\subsection{Molecular Mechanics Force Field} \label{sec:mm}
In this work, we follow the analytical forms in GAFF\cite{wangDevelopmentTestingGeneral2004} and OpenFF\cite{wangOpenForceField2024a}:
\begin{equation}
    E^\mathrm{MM} = E_{\mathrm{bonded}}^\mathrm{MM} + E_{\mathrm{non-bonded}}^\mathrm{MM}
\end{equation}
\begin{equation}
    \begin{split}
    E_{\mathrm{bonded}}^\mathrm{MM} =& E_{\mathrm{bond}}^\mathrm{MM} + E_{\mathrm{angle}}^\mathrm{MM} + E_{\mathrm{proper}}^\mathrm{MM} + E_{\mathrm{improper}}^\mathrm{MM} \\
    =& \sum_{\mathrm{bonds}} \frac{1}{2} k_{r, ij}(r_{ij}-r_{ij}^0)^2
    + \sum_{\mathrm{angles}} \frac{1}{2} k_{\theta, ijk}(\theta_{ijk}-\theta_{ijk}^0)^2 \\
    &+ \sum_{\mathrm{propers}} \sum_{n_\phi} k_{\phi, ijkl}^{n_\phi}\left[1+\cos(n_\phi\phi_{ijkl}-\phi_{ijkl}^{n_\phi,0})\right]\\
    &+ \sum_{\mathrm{impropers}} k_{\psi, ijkl}\left[1+\cos(2\psi_{ijkl}-\pi)\right]\\
    E_{\mathrm{non-bonded}}^\mathrm{MM} =& E_{\mathrm{vdW}}^\mathrm{MM}+E_{\mathrm{Coulomb}}^\mathrm{MM}\\
    =& \sum_{i<j}\left[ 4\epsilon_{ij}\left(\frac{\sigma_{ij}^{12}}{r_{ij}^{12}} - \frac{\sigma_{ij}^{6}}{r_{ij}^{6}}\right) + \frac{q_i q_j}{4\pi\epsilon_0 r_{ij}} \right]
    \end{split}
    \label{eq:mmff} 
\end{equation}
where the bond lengths $r$, angles $\theta$, proper torsion angles $\phi$ and improper torsion angles $\psi$ are rotationally invariant internal coordinates.
The force field parameters include bonded parameters (equilibrium values $r^0$, $\theta^0$ and $\phi^0$, and force constants $k_r$, $k_\theta$, $k_\phi$ and $k_\psi$), and non-bonded parameters (van der Waals (vdW) parameters $\sigma$ and $\epsilon$, and partial charges $q$).
A well-trained transferrable MMFF shall predict these parameters accurately for any given molecules.
In this article, ``torsion'' refers to proper torsion unless otherwise specified.
We also fix the phase angles $\phi_{ijkl}^{n_\phi,0}$ at 0 for odd $n_\phi$ and $\pi$ for even $n_\phi$, and therefore ensure that the torsional energy is independent of the order being $ijkl$ or $lkji$.

Force field parameters should adhere to several physical constraints:
(1) Force field parameters should be permutationally invariant, 
e.g., the force constant of bond $(i, j)$ should be equal to that of bond $(j, i)$.
(2) Force field parameters should be in accordance with chemical symmetries of molecules,
e.g., force constants of the two C-O bonds in a carboxyl group (-C([O-])=O) must be equal to each other, due to their chemical equivalency, even though they may have different bond orders assigned when written as a SMILES or SMARTS string. \cite{march1977advanced}
(3) Charge conservation should be guaranteed,
i.e., the summation of assigned partial charges of atoms in one molecule should be consistent with the molecule's net charge, which avoids net charge gain/loss for the molecule in the parameter determination.
These constraints are naturally satisfied in the traditional look-up table approaches,
which should also serve as essential guidelines when force field parameters are inferred by a machine-learning model.

Additionally, there are two key philosophies worth considering when building general small molecule MMFFs:
(1) Force field parameters should be dominated by local structures, so that the parameters trained from small molecules can be consistently transferred to similar structures in relatively large molecules. \cite{wangDevelopmentTestingGeneral2004,roosOPLS3eExtendingForce2019}
(2) Torsional energy profiles should be accurately captured, as the quality of torsion parameters significantly affects the conformational distribution of small molecules, thereby influencing the prediction of properties such as protein-ligand binding affinity. \cite{laheyBenchmarkingForceField2020,roosOPLS3eExtendingForce2019,hortonOpenForceField2022}
\subsection{Dataset Construction}

\subsubsection{Molecular fragments generation}
The curated dataset is mostly built from the ChEMBL database\cite{zdrazilChEMBLDatabase20232024}  with some additions from the ZINC20\cite{irwin2020zinc20} datasets to further enhance the diversity. 
From these datasets, a subset of molecules was initially selected by several criteria, including number of aromatic rings, polar surface area (PSA), quantitative estimate of drug-likeness (QED), element types, and hybridization types.
These selected molecules were then cleaved into fragments with less than 70 atoms using our in-house graph-expansion algorithm (details in supporting information), such that the local chemical environments were well-preserved. \cite{jinsong2024molecular,roosOPLS3eExtendingForce2019} 
In brief, this fragmentation algorithm traverses over each bond, angle, and non-ring torsion in a molecule, retains the relevant atoms and their conjugated partners, then trims the rest and caps the cleaved bonds.
Next, these fragments were expanded to various protonation states within a pKa range of 0.0 to 14.0, calculated by Epik 6.5, \cite{shelley2007epik} to cover most possible protonation states that might appear in aqueous solutions.
Finally, 2.4 million unique fragments were selected for QM calculations after deduplication.

\subsubsection{Quantum chemistry methods and workflow}

From the 2.4 million fragments, we created two QM datasets, namely the \textit{optimization dataset} and the \textit{torsion dataset}, at the B3LYP-D3(BJ)/DZVP level of theory.
The same QM method is also used by OpenFF to generate their training data. \cite{qiuDevelopmentBenchmarkingOpen2021,boothroydDevelopmentBenchmarkingOpen2023}
This method achieves a good balance between accuracy (relative to CCSD(T)/CBS) and computational cost in recent benchmarks. \cite{Kesharwani2015,Behara2024}
For molecular conformational PES, more advanced methods such as $\omega$B97M-V are significantly more expensive but only marginally more accurate.

The \textit{optimization dataset} was generated from the entire 2.4 million fragments. 
The 3D conformations of each fragment was initially generated by RDKit from its SMILES string, and then optimized using the geomeTRIC \cite{Wang2016} optimizer at the chosen QM level. 
The relaxed structure is verified where no accidental bond breaking or formation happened during the structural relaxation.
Finally, the Hessian matrix is calculated for each fragment using Q-Chem 6.1\cite{Epifanovsky2021}.
The hessian matrix is further verified by checking all the eigenvalues (except for six near-zero values corresponding to translational and rotational modes) to ensure that a true local minimum is captured.
More details about the workflow and the filtering criteria are provided in the supporting information.

The \textit{torsion dataset} consists of two subsets, non-ring torsions and in-ring torsions, curated separately, comprising 2.2 million and 1.0 million frames respectively.
For each non-ring torsion fragment, we first rotate the torsion angle in 15$^\circ$ increments from the optimized conformation, creating 24 initial frames that were then separately relaxed using the geomeTRIC optimizer with the rotated torsion angle constrained.
While the in-ring torsions were scanned using a sequential frame-by-frame approach, with an early stopping strategy if the conformational energy is more than 20 kcal/mol higher than the unconstrained relaxed conformation, since those high-energy regimes are irrelevant in room-temperature molecular dynamics simulations.
For both non-ring and in-ring subsets, the resultant conformers were also carefully filtered by the bond breaking or formation criteria.
Due to the high computational cost of torsion scan, these calculations were performed with GPU4PySCF\cite{li2024introducinggpuaccelerationpythonbasedsimulations,wu2024enhancinggpuaccelerationpythonbasedsimulations} for significant acceleration and cost reduction. 
The results were also verified to be consistent with Q-Chem.

\subsubsection{Diversity of Chemical Space}
Evaluating the diversity of chemical space composed of numerous molecules often involves representing molecules with descriptors or fingerprints followed by measuring of their distances. 
One widely used fingerprint is the Morgan fingerprint, which captures the local environment of each atom within a specified radius defined by the number of covalent bonds.
To emphasize the local structure of torsions, Morgan fingerprints were calculated with a radius of 2, focusing exclusively on the neighborhoods of the two central atoms in the selected torsion.
This torsional fingerprint was used to evaluate the diversity of different datasets, including SPICE, GEOM, and our \textit{torsion dataset}, as well as in the creation of a diverse torsion dataset (BDTorsion). 

\subsection{ByteFF Model}
\label{sec:model}
\begin{figure}[H]
\centering
\includegraphics[width=0.5\linewidth]{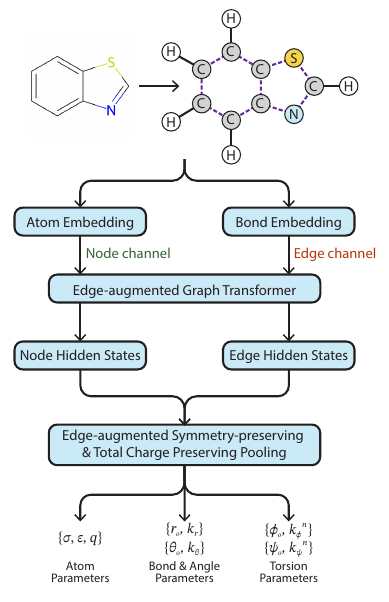}
\caption{\textbf{Model structure of ByteFF.} 
ByteFF predicts MMFF parameters in three steps. 
First, atom and bond features are extracted from the molecular graph and then projected into embeddings. 
Then, an edge-augmented graph transformer (EGT)\cite{hussainGlobalSelfAttentionReplacement2022} is used to synergize the edge embeddings with the node-based attention mechanism. 
Lastly, the output module derives force field parameters while preserving the molecular symmetry and total charge.
}
\label{fig:model_structure}
\end{figure}
ByteFF predicts both the bonded and non-bonded force field terms in one-pass.
It leverages the state-of-the-art Edge-augmented Graph Transformer (EGT) \cite{hussainGlobalSelfAttentionReplacement2022} as its backbone to achieve accuracy and transferability.
The output of EGT is further processed by a series of MLPs to derive force field parameters while preserving both the total charge and the physical symmetry of each force field term.
The ML model structure is shown in Fig.~\ref{fig:model_structure} and consists of three major parts: featurization, GNN, and output modules.
They are briefly described here, and more details are given in the supporting information.

\textbf{Featurization module}. 
For each molecule, we first convert the molecule structure into a graph, and detect chemically equivalent atoms and bonds with bond orders ignored.
Atomic features are then detected from this graph, including the element type, formal charge, ring connectivity and minimum ring size. 
Additionally, to better describe the chemical environment of each atom, bond features including bond order and whether the bond is in ring, are also extracted.
Replacing the discrete atom-typing rules used in conventional MMFFs, atom and bond features are projected into continuous vectorial embedding spaces, and concatenated into node and edge feature embeddings, respectively.
Importantly, embeddings of chemically equivalent atoms and bonds are averaged before passed to the next module, which helps to preserve the chemical symmetry of the molecule.

\textbf{GNN module}.
In the GNN module, a modified EGT architecture is employed to learn the chemical environment of molecules from both atom and bond features.
This architecture synergizes the node and edge features in the attention mechanism to efficiently capture the structural information of the molecule.
To ensure locality and speed up training and inference, the original global self-attention in EGT is modified such that attention is applied only to a local neighborhood of each node.

\textbf{Output module}. 
After a few layers of localized EGT, atom and bond hidden states generated by the GNN module are passed to the output module to predict each force field term individually.
For each bonded term, the corresponding output module is designed to preserve the necessary symmetry, respectively.
For the partial charge term, the output module is designed to preserve the total charge of the molecule following a bond-charge-correction (BCC) style postprocessing. 

Additionally, an ensemble of five models randomly initialized is trained for improved predictive performance and uncertainty quantification.
\subsection{Training Procedures}
\label{sec:train}
To efficiently use the curated \textit{optimization} and \textit{torsion datasets}, while enhancing the robustness and performance of the ByteFF model, we took an ingenious training strategy with three stages: pre-training, training, and fine-tuning.
The loss functions (details in supporting information) are carefully designed in each stage such that the improvement of the force field is properly reflected by the minimization of loss functions. 

In the pre-training stage, the non-bonded parameters and force constants of proper torsions were fitted to GAFF-2.2 with mean squared error (MSE) losses, while the equilibrium values of bonded parameters were fitted with energy-based loss functions, using the \textit{optimization dataset}.
Meanwhile, the force constants of bonded terms in Eq.~\ref{eq:mmff} were trained using a partial Hessian loss, evaluated by the mean absolute percentage error (MAPE) of Hessian blocks corresponding to bonds, angles and improper torsions, namely partial Hessian blocks.
It has been reported that the partial Hessian blocks can be used to derive accurate force constants for single molecules. \cite{seminario1996calculation,allen2018harmonic,wangPartialHessianFitting2016}
Our design of the differentiable partial Hessian loss enables fitting force constants in batches, combining training accuracy with computational efficiency.

Given the significance of torsion profiles in the quality of force fields, in the training stage, we incorporated the curated \textit{torsion dataset} to fit the force constants of proper torsions, while the other parameters were trained using \textit{optimization dataset} following the same manner in the pre-training stage.
Here, we employed the Boltzmann MSE loss functions \cite{dahlgrenCharacterizationBiarylTorsional2013} into an iterative optimization-and-training process to train the force constants of proper torsions using the \textit{torsion dataset}.
In each iteration, the QM-optimized geometries were refined by the force field, with the torsion angle constrained and atom positions restrained \cite{qiuDevelopmentBenchmarkingOpen2021}. 
The parameters were then trained on both the \textit{optimization} and \textit{torsion datasets}. 
As the force field's accuracy improved, the positional restraint force constant was gradually reduced.
Additionally, L1-norm regularization was applied to the force constants of proper torsions to restrain redundant degrees of freedom.
The combination of pre-training and training stages yields the ByteFF-gopt model.

In the final fine-tuning stage, we incorporated part of the training set in Espaloma-0.3.0\cite{takabaMachinelearnedMolecularMechanics2024}, named \textit{off-equilibrium dataset} in this work, to refine the force field parameters with the QM energy and forces.
Combining all the three stages (pre-training, training, and fine-tuning), the ByteFF-joint model was obtained.

The detailed settings of the three-stage training procedure are summarized in Table~\ref{tab:training_parameters}.

\begin{table}[ht]
    \centering
    \caption{Summary of training settings for different stages}
    \label{tab:training_parameters}
    \resizebox{\textwidth}{!}{
        \begin{tabular}{llll}
            \hline
            & \textbf{Pre-training} & \textbf{Training} & \textbf{Fine-tuning}\\ 
            \hline
            \multirow{2}{*}{\makecell[l]{Model parameters\\initialized}} & \multirow{2}{*}{\makecell[l]{randomly initialized\\parameters}} & \multirow{2}{*}{\makecell[l]{parameters from\\pre-training stage}} & \multirow{2}{*}{\makecell[l]{parameters from\\training stage}} \\
            &&& \\
            \hline
            \multirow{8}{*}{\makecell[l]{Datasets and\\fitting targets}} & 
            \multirow{8}{*}{\makecell[l]{\textit{optimization dataset}:\\- GAFF-2.2 $\sigma / \epsilon / q / k_{\phi}$ \\ - $E^\mathrm{MM}_\mathrm{bond}$ / $E^\mathrm{MM}_\mathrm{angle}$ / $E^\mathrm{MM}_\mathrm{improper}$ \\ - QM Hessian matrix}} &
            \multirow{8}{*}{\makecell[l]{\textit{optimization dataset}:\\- GAFF-2.2 $\sigma / \epsilon / q $ \\ - $E^\mathrm{MM}_\mathrm{bond}$ / $E^\mathrm{MM}_\mathrm{angle}$ / $E^\mathrm{MM}_\mathrm{improper}$ \\ -  Hessian matrix \\ \textit{torsion dataset}: \\ - QM energy \\ - L1-norm of $k_\phi$ }}&
            \multirow{8}{*}{\makecell[l]{ \textit{optimization dataset}:\\ same as training \\ \textit{torsion dataset}: \\ same as training \\ \textit{off-equilibrium}\\\textit{dataset}: \\ - QM energy/force}}\\
            &&& \\
            &&& \\
            &&& \\
            &&& \\
            &&& \\
            &&& \\
            &&& \\
            \hline
            \multirow{3}{*}{\makecell[l]{Optimization-and-\\training strategy}} & 
            \multirow{3}{*}{no force field optimization} & 
            \multirow{3}{*}{\makecell[l]{three optimization-and-train\\iterations on \textit{torsion dataset}}} & 
            \multirow{3}{*}{\makecell[l]{using coordinates from\\the last iteration of \\training stage}}\\
            &&& \\
            &&& \\
            \hline
            Optimizer & RAdam & RAdam & RAdam \\
            \hline
            Learning rate & $10^{-4}$ & $10^{-4}$ & $2 \times 10^{-5}$ \\
            \hline
            \multirow{5}{*}{Scheduler} &
            \multirow{5}{*}{\makecell[l]{ReduceLROnPlateau \\ factor: 0.2 \\ patience: 10 \\ threshold: $10^{-6}$}} &
            \multirow{5}{*}{\makecell[l]{ReduceLROnPlateau \\ factor: 0.2 \\ patience: 4 \\ threshold: $10^{-6}$}} &
            \multirow{5}{*}{\makecell[l]{ReduceLROnPlateau \\ factor: 0.2 \\ patience: 4 \\ threshold: $10^{-6}$}}\\
            &&& \\
            &&& \\
            &&& \\
            &&& \\
            \hline
            Early stop patience & 50 & 10 & 10 \\ 
            \hline
            Yield & - & ByteFF-gopt & ByteFF-joint \\
            \hline
        \end{tabular}
    }
\end{table}

\subsection{Benchmark Datasets and Metrics}
To be compatible with the Amber family of protein force fields, the non-bonded terms of ByteFF were trained to reproduce the AM1-BCC charges \cite{jakalian2000fast,jakalian2002fast} and GAFF-2.2 vdW parameters.
Therefore, in this work we focus on the intramolecular PES performance relative to QM references of ByteFF, and leave the further improvement of non-bonded terms as a future work.  

Three different types of benchmarks were used to provide a thorough assessment of ByteFF.
First, the large-scale industrial collaborative OpenFFBenchmark public dataset \cite{damoreCollaborativeAssessmentMolecular2022} was used to quantify how well the force field-relaxed geometries and energies reproduced the QM-relaxed references, without constraints.
Three different metrics were used, namely 
root-mean-square deviation of atomic positions (RMSD),
torsion fingerprint deviation (TFD) \cite{schulz-gaschTFDTorsionFingerprints2012},
and relative energy difference ($\Delta\Delta E$) defined by OpenFF team.\cite{10.12688/f1000research.27141.1}
Second, the crucial torsional energy profile performance was benchmarked on two different datasets, namely the publicly available TorsionNet500 dataset \cite{raiTorsionNetDeepNeural2022} and the BDTorsion dataset curated in-house.
For consistency, energies of conformations in TorsionNet500 \cite{raiTorsionNetDeepNeural2022} dataset were re-calculated with the B3LYP-D3(BJ)/DZVP level of theory.
On these datasets, the root mean squared error (RMSE) and Boltzmann-weighted RMSE of each torsional energy profile were calculated and compared.

Beyond the constrained (torsional energy) and unconstrained (OpenFFBenchmark) optimization benchmarks, the performance of ByteFF was further benchmarked on the QM references energies and forces calculated on the \textit{off-equilibrium dataset},
which includes subsets of the SPICE dataset\cite{eastmanSPICEDatasetDruglike2023} and the RNA Structure Atlas\cite{parlea2016rna}.
On this dataset, RMSE of energies and forces were calculated to quantify if the PES is properly captured beyond local minima.

Rigorous procedures were followed in the benchmarking of ByteFF models, which ensured no data leakage between the training and testing data.
More details of these benchmarks can be found in supporting information.

\section{Results and Discussion}
\subsection{Diversity of Various Datasets}
\begin{figure*}[h]
    \centering
    \includegraphics[width=0.85\linewidth]{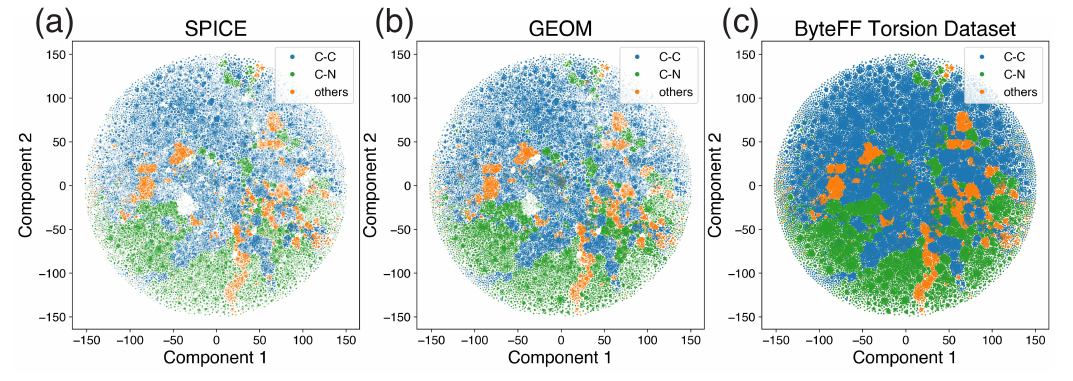}
    \caption{\textbf{t-SNE analysis of different datasets.} 
    The Morgan-based torsional fingerprint analysis results are illustrated using the t-SNE algorithm. 
    Every scatter dot corresponds to a torsion profile being analyzed in the corresponding dataset, which is colored by the element types of the two center atoms in the torsion.}
    \label{fig:t-SNE}
\end{figure*}

Using the Morgan-based torsional fingerprint defined above, we quantify the diversity of SPICE, GEOM and BDTorsion datasets in terms of torsional local environment.
Specifically, in all three datasets, only torsions with circular standard deviation \cite{mardia1972InStatistics} greater than 0.3 were included and used in fingerprint analysis.
The diversity of each dataset was then visualized using the standard t-SNE (t-Distributed Stochastic Neighbor Embedding) algorithm. 
In Fig.~\ref{fig:t-SNE}, the scatter dots are colored by the element types of the two center atoms (the rotatable bond) in each torsion.
The most frequent rotatable bond is C-C, followed by C-N rotatable bonds and other bonds including C-O, C-S, etc.
It is evident that our \textit{torsion dataset} excels in the diversity and comprehensiveness of its chemical space coverage, while the SPICE and GEOM datasets are more sparse.
In comparison, both SPICE and GEOM datasets show obvious vacancies, corresponding to the absence of certain chemical torsion patterns.
For example, the uncovered areas around (-45, -20) in Fig.~\ref{fig:t-SNE} (a) and (b) correspond to in-ring torsions of protonated pyridine and imidazole, as well as their derivatives, which are indispensable for the training of force fields for bio-organic molecules. 
Such an expansive coverage of the chemical space in the training dataset provides a solid foundation for the coverage and transferability of ByteFF.

\subsection{Torsional Potential Energy Surfaces}

To evaluate the accuracy of ByteFF in predicting torsional PES, and directly compare with the performance of other force fields, we performed comprehensive benchmarks using the TorsionNet500 and BDTorsion datasets.
TorsionNet500 \cite{raiTorsionNetDeepNeural2022} is a benchmark dataset consisting of 500 chemically diverse fragments relevant to biological and pharmaceutical applications.
The BDTorsion dataset is curated in this work.
It is divided into Non-Ring and In-Ring torsion subsets and each contains 1000 chemically diverse fragments.
For both TorsionNet500 and BDTorsion datasets, dozens of conformations are sampled by scanning torsion angle, then QM-relaxed, and tagged with QM energy labels for each fragment.
Here, two metrics were used to evaluate the accuracy of the predictions from various force fields with respect to the ground truth from QM: the RMSE and Boltzmann RMSE, with RMSE assessing the overall discrepancy while Boltzmann RMSE emphasizing the deviations near the energy minimum.
When calculating the Boltzmann RMSE, we used 2.0 kcal/mol for both cutoff and scaling factors.

\begin{figure}[H]
    \centering
    \includegraphics[width=0.85\linewidth]{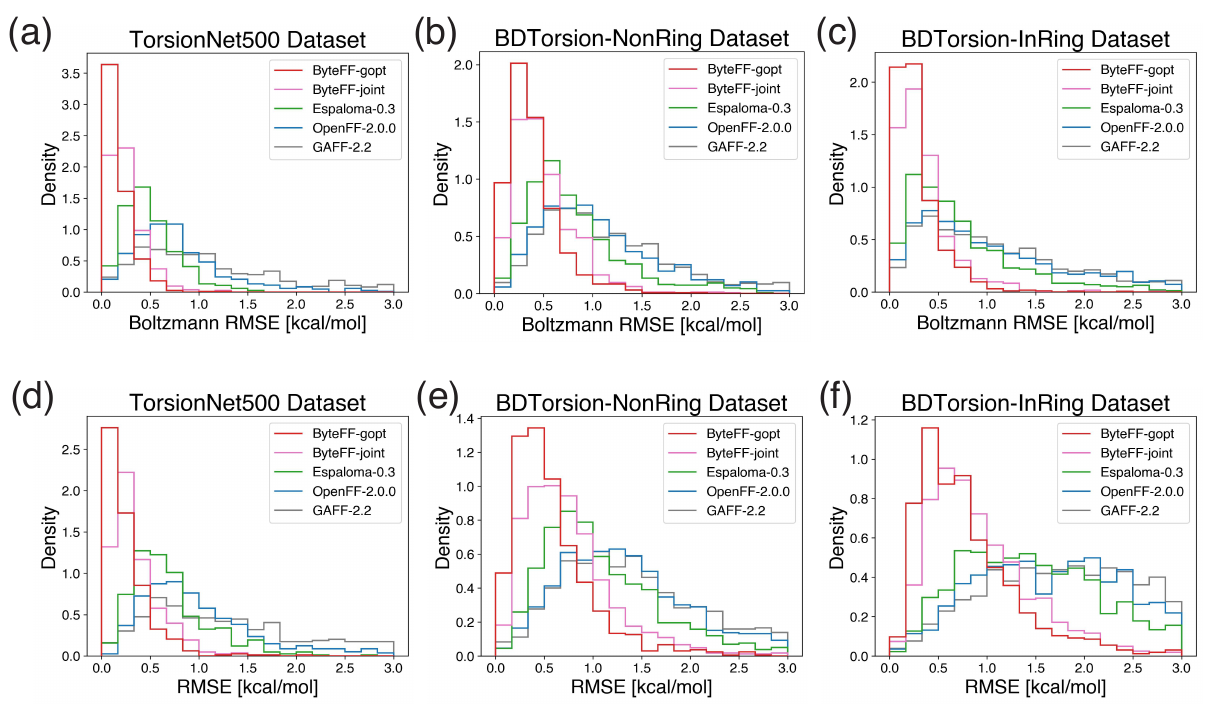}
    \caption{\textbf{Histograms of the discrepancy of torsional PES between predictions of QM and force fields.} As a comprehensive benchmark, two metrics including Boltzmann RMSE (a-c) and RMSE (d-f) are used to assess the accuracy of force field-predicted torsional energy profiles with respect to the QM results. Three datasets were included in this benchmark: TorsionNet500 (a \& d), BDTorsion-NonRing (b \& e), and BDTorsion-InRing (c \& f).} 
    \label{fig:torsion}
\end{figure}

As illustrated in Fig.~\ref{fig:torsion}, both ByteFF-gopt and ByteFF-joint significantly outperform competitors on all three benchmark datasets.
On all the three datasets, ByteFF-gopt exhibits the best performance on both metrics, with the highest population of predictions with lower RMSE and Boltzmann RMSE.
When tested on TorsionNet500, the Boltzmann RMSE of most predictions by ByteFF-gopt are within 0.5 kcal/mol, while none of the Boltzmann RMSE exceeds 1.33 kcal/mol, demonstrating the exceptional accuracy of ByteFF-gopt in terms of predicting torsional energy profiles.
On the more challenging BDTorsion datasets, both RMSE and Boltzmann RMSE are generally higher for all the force fields being tested.
Nevertheless, ByteFF-gopt and ByteFF-joint still outperformed competitors significantly, with most Boltzmann RMSE values below 1.0 kcal/mol (chemical accuracy).
In these torsion profile benchmarks, the performance of ByteFF-joint is compromised by the addition of off-equilibrium energies and forces training data.
This is inevitable due to the very limited fixed functional form of MMFFs.

\begin{figure}[h]
    \centering
    \includegraphics[width=0.5\linewidth]{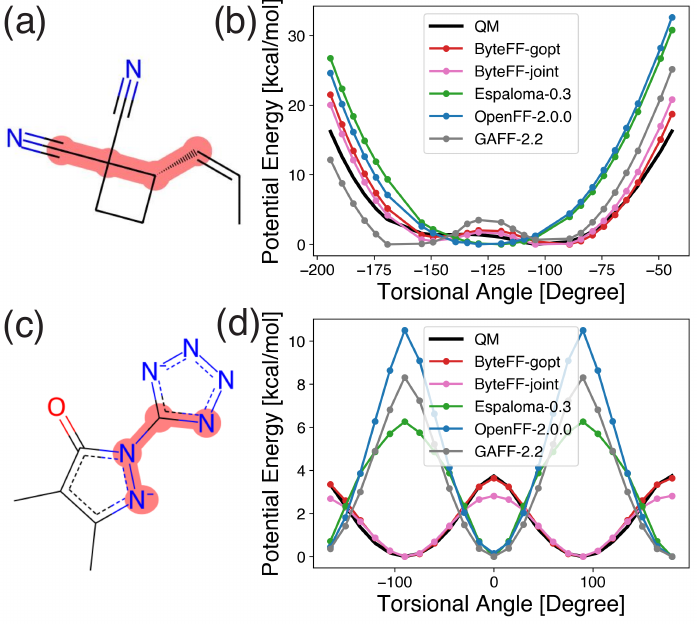}
    \caption{\textbf{Example of the in-ring and non-ring torsion prediction accuracy of various force fields.} 
    As examples to show the accuracy of ByteFF models in predicting torsional energy profiles, an in-ring (a-b) and a non-ring (c-d) example molecule are provided.
    The torsional energy profiles predicted by various force fields are compared with the QM references and shown for each example molecule.}
    \label{fig:InRing_example}
\end{figure}

One key feature of organic molecules is the presence of a great variety of rings\cite{shearerRingsClinicalTrials2022}, and in-ring torsion parameters have crucial impact on the conformational PES of rings.
However, this is much less discussed in the development of MMFFs.
In this work, we carefully constructed the in-ring torsion scan training dataset to improve the performance of ByteFF on in-ring torsions.
As shown in Fig.~\ref{fig:torsion} (c) and (f), both ByteFF-gopt and ByteFF-joint significantly outperform competitors, with both RMSE and Boltzmann RMSE mostly better than chemical accuracy, respectively. 
As an example to illustrate the accuracy of ByteFF-gopt and ByteFF-joint in predicting in-ring torsional energy profile, we present a molecule from the BDTorsion-InRing dataset that contains a four-membered aliphatic ring (Fig.~\ref{fig:InRing_example} (a)).
On this molecule, we calculated the torsional energy profile for the in-ring torsion atoms highlighted in Fig.~\ref{fig:InRing_example} (a), and compared PES predictions from various force fields with the QM references (Fig.~\ref{fig:InRing_example} (b)).
It is evident that both Espaloma-0.3.0 (green) and OpenFF-2.0.0 (blue) failed to predict the energy landscape in the vicinity of the local minimum, which could lead to incorrect predictions of ring conformations in geometric optimization or MD simulations.
Though GAFF-2.2 captured the approximate shape of this torsional energy landscape, it overestimated the barrier near -130$^\circ$ while significantly underestimated the energy near  -160$^\circ$, which led to significantly incorrect shifts of the predicted location of global energy minimum.
In contrast, both ByteFF-gopt and ByteFF-joint predictions align well with QM references, which not only captured the shape of the energy landscape, but also accurately predicted the positions and energy differences of local minima. 
In addition to accurately predicting in-ring torsional energy profiles, ByteFF models also excel in the prediction of non-ring torsions, as illustrated with the example molecule in Fig.~\ref{fig:InRing_example} (c).
As shown in Fig.~\ref{fig:InRing_example} (d), ByteFF-gopt and ByteFF-joint successfully predicted the torsional energy landscape of the highlighted non-ring torsion, which is highly consistent with the QM reference, while other force fields failed to predict both the positions of energy minima and the height of the energy barrier.
More example molecules are provided in Fig. S2-S3 including both in-ring and non-ring torsions, where ByteFF models predicted the torsional energy profiles with exceptional accuracy against the QM reference, while other force fields failed to achieve.

The effectiveness and accuracy of ByteFF-gopt and ByteFF-joint stems from the synergy of two components: comprehensive in-house training datasets and carefully tweaked training strategy.
The iterative optimization-and-training strategy allows the model to accurately capture the energy landscape along torsional degrees of freedom, minimizing interference from complex interactions such as bond-torsion coupling that is beyond the limited description of the chosen functional form.
The predictions are further enhanced by the simple but effective model-ensemble averages.

\subsection{Equilibrium and Off-Equilibrium Conformations}

In addition to predict the torsional PES of molecules, the OpenFFBenchmark\cite{damoreCollaborativeAssessmentMolecular2022} dataset is used to benchmark the performance of various force fields in terms of predicting unconstrained equilibrium conformations and associated energies.

\begin{figure}[H]
    \includegraphics[width=\linewidth]{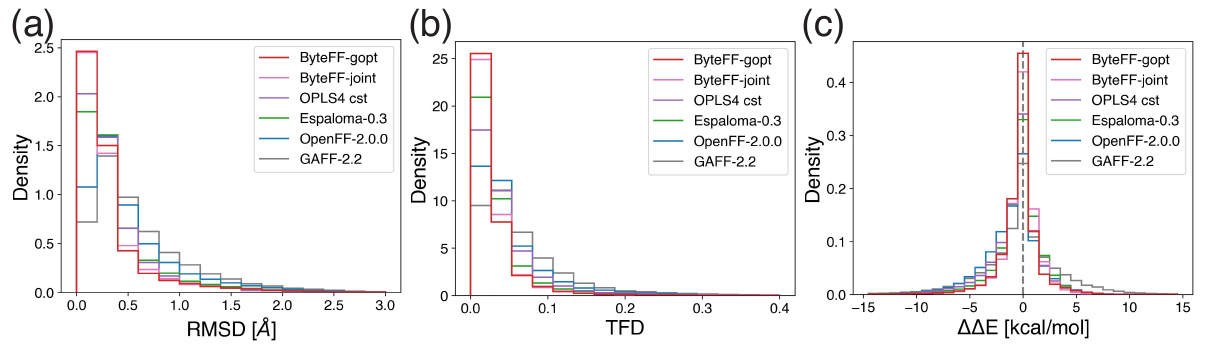}
    \caption{\textbf{Histograms of different metrics on OpenFFBenchmark dataset.} 
    The accuracy of force field-relaxed geometry relative to QM-relaxed references is quantified with (a) RMSD and (b) Torsion Fingerprint Deviation (TFD) scores. 
    The energetic accuracy is quantified by (c) $\Delta\Delta E$ distributions. 
    All benchmark results for ``OPLS4 cst'', and the detailed protocols to calculate the benchmark results are obtained from ref \citenum{damoreCollaborativeAssessmentMolecular2022}.}
    \label{fig:openffbenchmark}
\end{figure}

As shown in Fig.~\ref{fig:openffbenchmark} (a) and (b), the distribution of ByteFF-gopt and ByteFF-joint in terms of RMSD and TFD are both concentrated near zero, with higher peak values near zero and narrower distributions than other force fields, demonstrating superior consistencies with QM-optimized molecular conformations.
Particularly, both ByteFF-gopt and ByteFF-joint significantly outperform the ``OPLS4 cst'' model (OPLS4 with refined parameters using the FFBuilder tool for each molecule individually), which is considered the state-of-the-art MMFF in current reports.
Additionally, we calculated the relative energy differences ($\Delta\Delta E$) to assess energetic agreement between force field-optimized conformations and the QM-optimized ones (Fig.~\ref{fig:openffbenchmark} (c)). 
Both ByteFF-gopt and ByteFF-joint exhibit sharp peaks around zero in their $\Delta\Delta E$ distribution, significantly outperform all other force fields, indicating that both of them reproduce correctly the QM relative energies between conformers.
The accurate predictions of equilibrium conformations ensures the correct conformations being sampled in MD simulations, while the precise prediction of $\Delta\Delta E$ guarantees the Boltzmann distribution in the conformational space is properly reproduced.

\begin{figure}[H]
    \centering
    \includegraphics[width=0.8\linewidth]{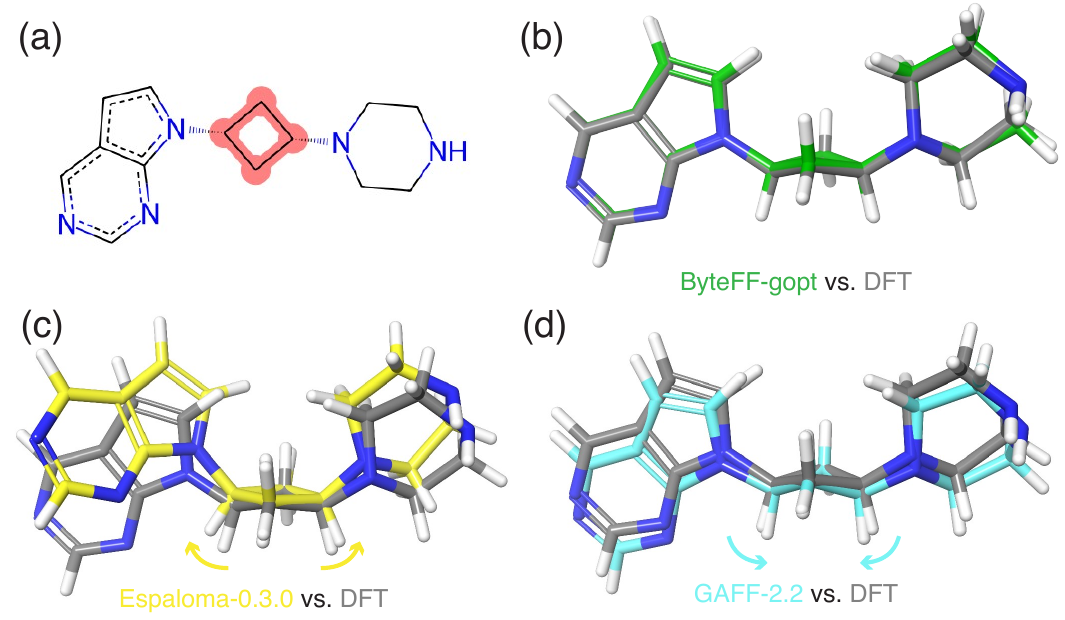}
    \caption{\textbf{Example of the equilibrium conformation predictions of molecules.} 
    (a) 2D representation of the example molecule. 
    (b-d) The equilibrium conformations predicted by ByteFF-gopt (green), Espaloma-0.3.0 (yellow), and GAFF-2.2 (cyan), superimposed to the equilibrium conformations predicted by QM (grey).}
    \label{fig:equi_conf}
\end{figure}

Fig.~\ref{fig:equi_conf} illustrates a more detailed example of the performance of reproducing equilibrium geometries for various force fields.
The QM-relaxed reference geometry shows that the four-membered ring takes a ``butterfly'' conformation due to significant torsional strain.
The predictions relaxed using various force fields, including ByteFF-gopt (green), Espaloma-0.3.0 (yellow), and GAFF-2.2 (cyan), are superimposed on the QM reference.
It is evident that Espaloma-0.3.0 (yellow) predicts the four-member ring as a planer conformation, while GAFF-2.2 (cyan) overestimates the bending of the ring.
Being at the center of the molecule, a minor misprediction of the four-member ring conformation results in more significant geometric deviations on the two ends of the molecule, leading to overall RMSD values of 0.9 \AA\ for Espaloma-0.3.0 and 0.4 \AA\ for GAFF-2.2, respectively.
Notably, the geometry relaxed by ByteFF-gopt (green) achieved superior alignment with the QM prediction with an RMSD of 0.2 \AA.
With increasing ring sizes and increasing number of neighboring rings, the complexity of the conformational space expands, making precise predictions of in-ring torsion profiles increasingly critical.

In addition to the equilibrium conformations, it is equally important for the MD simulations to accurately predict the off-equilibrium conformations.
We used several datasets to comprehensively assess the prediction accuracy of both energy and force predictions from various force fields, quantified by RMSE relative to QM references (Table~\ref{tab:off_equi}). 
Among all five subsets, ByteFF-gopt achieved higher accuracy than both GAFF-2.2 and OpenFF-2.0, but less accurate than Espaloma-0.3.0.
Importantly, ByteFF-gopt achieved this accuracy with relaxed geometry, energy, and partial hessian, without using any force labels in the training process.
With additional off-equilibrium energy and forces training data, we further trained ByteFF-joint to improve the force prediction capabilities of ByteFF-gopt.
Remarkably, after fine-tuning with just around 10k data in the \textit{off-equilibrium dataset}—a significantly smaller dataset than that used by Espaloma-0.3.0—the performance of ByteFF-joint surpassed that of its competitors. 

As a result, ByteFF-joint achieves state-of-the-art performance in almost all subsets, except for the SPICE-Pubchem dataset, where ByteFF-joint shows marginally higher force RMSE compared to Espaloma-0.3.0.
As discussed earlier, it is inherently challenging of an MMFF to perfectly replicate the QM PES. 
There is a trade-off between predicting the PES of relaxed-geometry and off-equilibrium conformations, and we focused more on the relaxed-geometry performances during the training process, 
since this is more relevant to the sampling of local minima in molecular dynamics simulations and is more stressed in the latest large-scale collaborative industrial benchmark\cite{damoreCollaborativeAssessmentMolecular2022}. 
Despite this, ByteFF-joint achieves overall state-of-the-art performance in all benchmarks considered in this study.

\begin{table}[ht]
\centering
\caption{Comparison of energy RMSE (in kcal/mol) and force RMSE (in kcal/mol/\AA) for different datasets and force fields.}
\label{tab:off_equi}
\resizebox{\textwidth}{!}{
    \begin{tabular}{llcccccc}
    \hline
    \textbf{Dataset} & & \textbf{GAFF-2.2} & \textbf{OpenFF-2.0} & \textbf{Espaloma-0.3} & \textbf{ByteFF-gopt} & \textbf{ByteFF-joint} \\
    \hline
    \multirow{2}{*}{SPICE-Pubchem} & \textit{Energy} & 4.8 & 4.4 & \textbf{2.3} & 3.4 & \textbf{2.3} \\
    & \textit{Force} & 14.1 & 14.0 & \textbf{6.5} & 9.9 & 6.7 \\
    \multirow{2}{*}{SPICE-DES-Monomers} & \textit{Energy} & 2.7 & 2.7 & 1.4 & 1.9 & \textbf{1.3} \\
    & \textit{Force} & 10.7 & 12.5 & 6.0 & 8.0 & \textbf{5.4} \\
    \multirow{2}{*}{SPICE-Dipeptide} & \textit{Energy} & 5.2 & 4.5& 3.3 & 4.0 & \textbf{2.3} \\
    & \textit{Force} & 12.5 & 12.4 & 8.1 & 9.9 & \textbf{5.3} \\
    \multirow{2}{*}{RNA-Diverse} & \textit{Energy} & 6.7 & 5.4 & 3.9 & 4.6 & \textbf{3.5} \\
    & \textit{Force} & 15.8 & 18.4 & 4.5 & 9.7 & \textbf{3.7} \\
    \multirow{2}{*}{RNA-Trinucleotide} & \textit{Energy} & 6.2& 6.0 & 3.8 & 4.4 & \textbf{3.4} \\
    & \textit{Force} & 16.5 & 18.7 & 4.4 & 10.0 & \textbf{3.6} \\
    \hline
    \end{tabular}
}
\end{table}

\subsection{Limitations}

Despite the excellent performance of ByteFF compared to other force fields, several limitations still exist.
As an MMFF with a minimal analytical functional, ByteFF inherently lacks the capability of modeling more complicated effects in intramolecular PES, such as bond-torsion and angle-torsion couplings.
Moreover, while intramolecular parameters are carefully trained against QM references, non-bonded parameters are directly inherited from GAFF-2.2 to ensure Amber compatibility, leaving room for further refinement. 
Addressing these limitations presents a path forward for the further improvements of ByteFF.

\section{Conclusion}
In this work, we developed ByteFF, an Amber-compatible MMFF with broad chemical space coverage and state-of-the-art accuracy. 
Through Morgan-based torsional fingerprints and t-SNE analysis, we demonstrated that our dataset covers superior chemical space.
Based on this strong data foundation, ByteFF leverages an edge-augmented, symmetry-preserving GNN model to predict MMFF parameters.
We also carefully designed our training strategy, enhanced by techniques such as partial Hessian fitting, an iterative optimization-and-training scheme, and effective model ensembling for prediction improvements and uncertainty quantification. 
Benefitting from the comprehensive datasets and exquisitely designed training strategy, ByteFF outperforms existing MMFFs across multiple benchmarks, delivering accurate predictions of torsional energy profiles, equilibrium conformations, and off-equilibrium energies and forces. 
We believe ByteFF will significantly benefit biomolecular simulations for various purposes.

\begin{acknowledgement}
\end{acknowledgement}

\bibliography{byteff}

\end{document}